\documentclass[review]{elsarticle}

\usepackage{lineno,hyperref}
\modulolinenumbers[5]
\usepackage{amsmath}
\usepackage{multirow}

\journal{Pattern Recognition Letters}

\begin{document}

\begin{frontmatter}

\title{Preserving Semantic Consistency in Unsupervised Domain Adaptation Using Generative Adversarial Networks}





\author{Mohammad Mahfujur Rahman \corref{cor1}} 
\cortext[cor1]{Corresponding author: 
  Tel.: +61416453032;}  
\ead{m27.rahman@qut.edu.au}
\author{Clinton Fookes}
\author{Sridha Sridharan}

\address{Signal Processing, Artificial Intelligence and Vision Technologies (SAIVT), Queensland University of Technology (QUT), Brisbane, QLD 4000, Australia}

\begin{abstract}
Unsupervised domain adaptation seeks to mitigate the distribution discrepancy between source and target domains, given labeled samples of the source domain and unlabeled samples of the target domain. Generative adversarial networks (GANs) have demonstrated significant improvement in domain adaptation by producing images which are domain specific for training. However, most of the existing GAN based techniques for unsupervised domain adaptation do not consider semantic information during domain matching, hence these methods degrade the performance when the source and target domain data are semantically different. In this paper, we propose an end-to-end novel semantic consistent generative adversarial network (SCGAN). This network can achieve source to target domain matching by capturing semantic information at the feature level and producing images for unsupervised domain adaptation from both the source and the target domains. We demonstrate the robustness of our proposed method which exceeds the state-of-the-art performance in unsupervised domain adaptation settings by performing experiments on digit and object classification tasks.
\end{abstract}


\end{frontmatter}


\section{Introduction}
\label{sec:intro}
A common requirement for the success of deep learning approaches is the availability of large amounts of labeled training data. These methods are generally trained on massive amounts of labeled training data and tested on the data that have the same distribution as the training data. When the distributions of the training and test data are different, the performance of the deep learning techniques degrades. To improve the performance in such situations, one needs to collect large amounts of labeled training data with the same distribution as the test data and this is a time consuming and expensive procedure. To address this problem one can use transfer learning where a learned model for one task is reused as the starting point for another task. We can reduce the expense of accumulating large sets of labeled data for training by transferring the information obtained from diverse but similar domains. Domain adaptation (DA) is a special type of transfer learning where the source data (training) and target data (testing) have a different distribution but share the same tasks.



Early DA techniques used hand-crafted features and were known as shallow DA methods \cite{6247911}. Deep neural networks extract more transferable and domain-invariant features. Thus, later efforts were motivated to extend shallow DA approaches to deep neural network based DA approaches \cite{DBLP:conf/icml/LongC0J15,pmlr-v37-ganin15,Rahman2020,RAHMAN2020107124}. Recently, generative adversarial networks (GANs) introduced by \cite{NIPS2014_5423} have been shown to be successful in the area of unsupervised domain adaptation (UDA). The GAN is used for DA for transforming the images from one domain to another. In \cite{Hu_2018_CVPR}, image translation and extracting domain invariant information is achieved using only a single encoder and a conditioned domain indicator code. In \cite{rucvpr}, domain alteration is achieved by using two encoders and two discriminators. In \cite{Gen2Adapt,rahman2019multi} the network produces source like images from both the source and target domains to minimize the discrepancy. However, the existing GAN based DA approaches \cite{Hu_2018_CVPR,Gen2Adapt,rucvpr} fail to preserve the semantic information across the source and target domains. As a result, the performance is compromised when the source and target domains are vastly different. 

\begin{figure*}
\begin{center}
\includegraphics[width=1.0\linewidth]{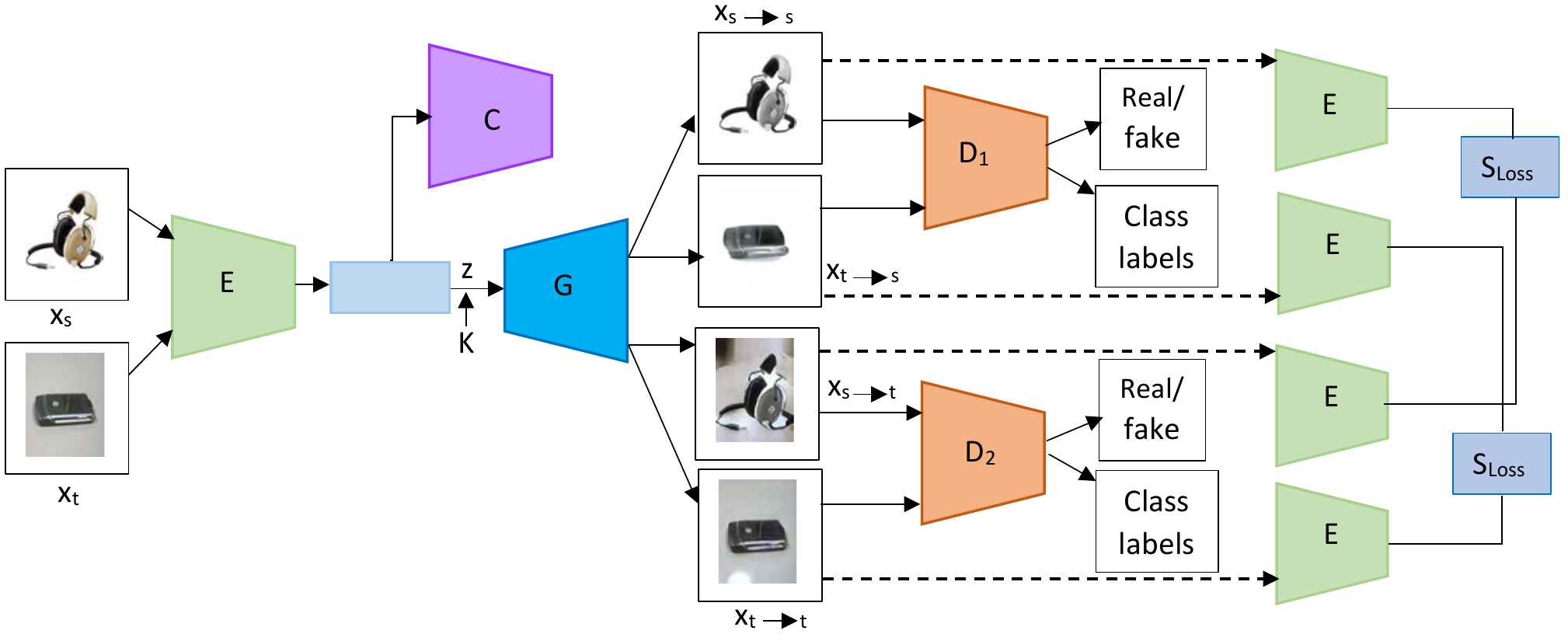}

\end{center}
   \caption{Illustration of our proposed approach. The proposed architecture comprises four main sections: an encoder $E$, a generator $G$, a classifier $C$ and two discriminators, $D_1$ for the source domain and $D_2$ for the target domain. The encoder embeds the images either from the source or target domain to extract latent embeddings. The generator decodes the latent embeddings from the encoder with a conditioned domain key. The generated images are also fed into the encoder to capture semantic information. The discriminators differentiate the real and generated images for both source and target domains.}
\label{fig:architecture}
\end{figure*}


To overcome the above-mentioned problem, we propose a semantic consistent generative adversarial network (SCGAN) for UDA to preserve the semantic information for the source and target domains. Although Hoffman et al. \cite{pmlr-v80-hoffman18a} proposed semantic consistency loss in DA task, the formulation of our proposed semantic consistency loss is significantly different. In \cite{pmlr-v80-hoffman18a}, the labels of the source data are used as it formulates the semantic consistency loss using cross-entropy loss; however, we formulate the semantic consistent loss using $L_1$ distance without using labels of the source data. Furthermore, \cite{pmlr-v80-hoffman18a} is not an end-to-end method for image classification task. In one step, it generates images and in another step, the generated images are used in the DA network to classify the images. An end-to-end deep domain adaptation strategy is able to tackle the domain bias through the incorporation of feature  extraction and domain alignment into a united architecture. Since the feature extraction module receives the feedback from the domain alignment module, an end-to-end deep domain adaptation strategy will perform better when compared to the not end-to-end domain adaption approaches. We show that our proposed approach, which preserves semantic consistency in domain transfer, has superior performance over the approach proposed by Hoffman et al 2018 which uses a semantic constancy loss. Our  approach also exceeds the performance of other state of the art approaches for DA.

Our proposed method is an end-to-end DA approach for the image classification tasks. We adopt pseudo labeling of unlabeled target samples that attempt to minimise the domain discrepancy. To achieve highly confident pseudo-labeled target data which is mostly correct, the model is pre-trained with labeled source data. The source and target data are passed through a common encoder during the training phase to obtain a latent representation. The generator then decodes the latent representation with the help of a domain key that ensures the domain alteration. The generated images are supposed to look like source or target styled images, hence we use two discriminators to identify the real or fake images from the source and target domains. Moreover, the discriminators act as a multi-class classifier to assure the gradient signals backpropagated by the discriminator for the target data corresponding to the respective classes. The generated images are passed through the encoder again to obtain the semantic representation of the source and target domains. The classifier which is used to predict the labels of the target data is developed on the latent representation extracted from the encoder. Our proposed approach achieves state-of-the-art performance on digit and object classification tasks in the UDA setting. The contributions of this work are summarised as follows:
\vspace{-1mm}
\begin{itemize}
\item A semantic consistent generative adversarial network (SCGAN) is proposed for UDA where the labeled source data and unlabeled target data are used during training, and the unlabeled data is used during testing.
\vspace{-1mm}
\item The proposed SCGAN preserves the semantic information across the source and target domains to improve the performance when domains are vastly different.
\vspace{-1mm}
\item SCGAN achieves superior results on object and digit classification tasks compared to the state-of-the-art.
\end{itemize}
\vspace{-5mm}

\section{Related Works}
\label{sec:related}
Adversarial learning is promisingly utilized in domain adaptation. Almost all the adversarial learning based domain adaptation approaches \cite{pmlr-v37-ganin15,8099799,pei2018multi} follow the concept from a generative adversarial network (GAN) \cite{NIPS2014_5423}. A discriminator is trained in such methods to determine whether the sampled feature derives from the source domain or target domain. By contrast, the feature extractor is trained to deceive the discriminator. Adversarial DA approaches address the domain shift problem in the feature-space whereas GAN based DA approaches \cite{NIPS2016_6544,pmlr-v80-hoffman18a,Gen2Adapt,Hong_2018_CVPR,Hu_2018_CVPR,gong2018causal,gong2019dlow,tran2019gotta,russo2018source,hu2018duplex,sankaranarayanan2018generate,murez2018image,hong2018conditional,lv2019targan} reduce the domain shift between the source and target data in the pixel-space by altering the source data to the style of a target domain.

Liu et al. \cite{NIPS2016_6544} developed coupled generative adversarial networks (CoGAN) to learn a shared distribution of the source and target domains utilizing two classifiers for two domains. In this approach, the predicting functions are adjusted to enable the source classifier to identify the target data accurately. Another GAN-based DA approach is introduced in \cite{Gen2Adapt} where the network generates source-styled images from the source and target embeddings. In \cite{Hong_2018_CVPR}, a conditional GAN is proposed for UDA in the context of a segmentation task. In \cite{Hu_2018_CVPR}, a duplex discriminator based GAN is used to mitigate the domain shift problem. 

Generative domain adaptation networks (G-DAN) is proposed in \cite{gong2018causal} which is capable to generate new domains by modifying the source and target distributions. Cycada \cite{pmlr-v80-hoffman18a} performed domain adaptation by translating source images into target styled images inspiring from the image-to-image translating method \cite{zhu2017unpaired}. Fengmao et al. \cite{lv2019targan} investigated the target data with their corresponding labels by disentangling the class code from the latent variables of the generator for target through the cooperation of the high mutual information and weight sharing process. Murez et al. \cite{murez2018image} combined cycle consistency, domain specific reconstruction and domain invariant feature extraction to transfer the source images into the target images and vice versa. Sankaranarayanan et al. \cite{sankaranarayanan2018generate} proposed a generative network aiming to transfer the target images into source like images. In \cite{russo2018source}, two generators are used for image translation from the source domain to target domain and vice-versa. However, the existing GAN based DA approaches fail to capture effectively the semantic information across the domains during domain translation. To mitigate this problem, the generated images are fed into an encoder to capture the semantic information in our proposed approach.

\vspace{-4mm}

\section{Proposed Method}
\vspace{-3mm}

\subsection{Problem Setup} 
In this section, we provide a brief illustration of the proposed SCGAN on UDA. We commence with a conventional representation of the domain adaptation problem. A domain \( \mathcal{D} \) is denoted by a joint distribution $P(X, Y)$ defined on  \( \mathcal{X \times Y} \) where \( \mathcal{X} \) represents the feature space and \( \mathcal{Y} \) represents the label space. We assume that we have access to labeled data $X_s$ = $\{x_s^i, y_s^i\}$ from the source domain $D_S$, and unlabeled data $X_t$ = $\{x_t^i\}$ from the target domain $D_T$. The source and target domain follow a different distribution; however, both domains share the same set of classes or categories. The aim of UDA is to seek a classifying function $F:$ $X$ $ \rightarrow $ $Y$ which is able to classify $X_t$ to the corresponding labels $Y_t$ given $X_s$ = $\{x_s^i, y_s^i\}$ and $X_t$ = $\{x_t^i\}$ during training as the input.


\vspace{-2mm}
\subsection{Semantic Consistent Generative Adversarial Network (SCGAN)}
\vspace{-1mm}

The aim of our proposed SCGAN which is composed of one encoder, one generator, one classifier, and two discriminators is to extract a domain invariant feature representation and provide domain alteration. Figure \ref{fig:architecture} shows an overview of our proposed approach. The encoder denoted as $E$ learns a latent embedding $E:$ $X$ $\rightarrow$ $Z$ from the input images from either the source or target domain; furthermore, the classification network, denoted as C, learns a prediction function $C:$ $Z$ $\rightarrow$ $Y$. As the target data is unlabeled, the classifying network has access to the labels of the source data only. The encoder learns the domain shift between $D_S$ and $D_T$ during extracting latent representation from the target data during training. The generator $G$ generates source and target styled images by decoding the extracted latent representation by the encoder with conditioned by a domain key using a reconstruction loss. As the reconstruction loss is used in the pixel-level, it fails to capture semantic features shared across the source and target domain which leads to a drop of performance when the source and target domains are vastly different. 

To capture the semantic features of the source and target domain, all the generated images are passed through an encoder and the semantic information is preserved by a semantic consistent loss. The tasks of the discriminators is to differentiate the real images from generated images and predict the class of the real images which compel the latent features to be domain specific and categorical knowledge is preserved. The labels of both source-styled from the source domain and target-styled from the source domain are available which can be directly used in the discriminators. However, the label of the source-styled from the target domain and target-styled from the target domain are not able to be obtained, so the pseudo categorical labels anticipated and assigned from the classifier are utilized by training the classifier with the source data. After achieving the convergence, the learned network provides annotation of the generated source-styled from target data and target-styled from target data. The discriminators ($D_1$) and ($D_2$) map the real image or the generated image into two distributions: the probability of the input being real, which is modeled as a binary classifier and the class probability distribution of the input, which is modeled as a $N_c$-way classifier. 

Now, we will briefly formalize the illustration above. At first, the encoder receives an input image $x_i$ either from the source domain or target domain and extracts a latent representation, $z = E(x_i)$. The classifier is built on the latent representation taking as input the latent representation which is generated by $E$ and predicts a multi-class distribution. The latent representation from the source and target domain can be denoted as $z_s$ and $z_t$ respectively. The latent representation $z$ is supposed to be domain specific. The generator $G$ takes the latent space representation as input from the feature extractor and produces images with a condition and $G$ can be represented as,
\begin{equation}
x_k = G (z, k), z \in z_s \cup z_t,
\end{equation}

\noindent where the domain key $k \in \{k_s, k_t \}$ is a one-hot vector used to determine the domain. 

The generator $G$ generates four types of images from two input domains. The generated images appear as either source-styled or target-styled images. For example, the source-styled image generated from the source data can be expressed as, 
\begin{equation}
x_{s \rightarrow s}  = G(E(x_s),k_s)  = G (z_s, k_s),
\end{equation}
where $x_{s \rightarrow s}$ is the source-styled image generated from source data.

Similarly, the target-styled image ($x_{s \rightarrow t}$) can be generated from source  data, the source-styled image ($x_{t \rightarrow s}$) can be generated from target data and the target-styled image ($x_{t \rightarrow t}$) can be generated from the target data. It is noted that the encoder and generator can be any type of deep learning network. When the source domain image $x_s$ is translated into source domain styled image $x_{s \rightarrow s}$, it is expected that the generated image should be the same as itself $x_s$. Similarly, when the target domain image $x_t$ is translated into target styled image, the generated image should be the same as $x_t$. Besides, the target-styled image from the source domain $x_{s \rightarrow t}$ should look like the real target domain and the source-styled image generated from the target domain $x_{t \rightarrow s}$ should look like real source domain.

Domain alteration can be achieved by the reconstruction loss and domain key. However, the reconstruction loss fails to capture the semantic information across the source and target domains. To capture the semantic information across the domains at the feature level, the generated images are passed through the encoder. For example, the encoder produces the latent representation from the source-styled generated images from source data as follows,
\begin{equation}
z_{s \rightarrow s} = E (x_{s \rightarrow s}),
\end{equation}

Similarly, the encoder produces the latent representation from the other source and target-styled images that are generated from the source and target data as $z_{s \rightarrow t}$, $z_{t \rightarrow s}$ and $z_{t \rightarrow t}$.

Finally, the objective function of the encoder and generator can be represented as,
\begin{align}
L_G = \underset{E, G}{min} \big(\sum_{x_s \in, D_S}(J (D_2 (x_{s \rightarrow t}), y_{s \rightarrow t}) + \alpha l_1(x_{s \rightarrow s}, x_{s}) +  \beta l_1(z_{s \rightarrow s}, z_{s \rightarrow t}) + \nonumber	\\ \sum_{x_t \in D_T}(J (D_1 (x_{t \rightarrow s}), y_{t \rightarrow s})  +  \alpha l_1( x_{t \rightarrow t}, x_{t})  + \beta l_1(z_{t\rightarrow t}, z_{t \rightarrow s}) \big),
\end{align}

\noindent where $J(.,.)$ is the cross entropy loss, $D_1$ and $D_2$ are two discriminators, $\alpha$ and $\beta$ are the hyper parameters for the reconstructed loss and semantic consistent loss respectively. $l_1$ is the Manhattan distance between the latent representations of the generated images.

The discriminators not only differentiate the real and generated images but also categorise the real source and target images based on the true labels of the source data and pseudo labels of the target data. The discriminator $D_1$ for the source domain aims to differentiate the real source image $x_s$ and the generated target styled image from source $x_{t \rightarrow s}$. Moreover, $D_1$ categorize the $x_s$ and $x_{t \rightarrow s}$ using true labels for source data and pseudo labels for source styled image from target domain. Similarly, $D_2$ for target domain aims to differentiate real and generated images as well as categorize the source-styled image from target domain and target-styled image from source domain. As the target-styled image from the source domain remains the same category as real source data, we use the true labels for the target-styled image from source domain and pseudo labels for the target-styled image from target data due to unavailability of the labels of target data.  

In summary, the objective function of the two discriminators $D_1$ and $D_2$ can be formulated as,
\begin{align}
L_D = \underset{D_1, D_2}{min} \big(\sum_{x_s \in, D_S}(J (D_1 (x_{s}), y_{s}) + J (D_2 (x_{s \rightarrow t}), y_{s \rightarrow t})  +  \nonumber	\\ \sum_{x_t \in, D_T}(J (D_1 (x_{t \rightarrow s}), y_{t \rightarrow s}) +  J (D_2 (x_{t}), y_{t})\big).
\end{align}
\subsection{Classifier}
A classifier $C$ is used on the latent embeddings that are extracted from the encoder for the classification network. The objective function of the classifier for the source and target data can be represented as follows,
\begin{equation}
L_c = \underset{C} {min} \big(\sum_{x_s \in D_S} J (z_s, y_s) + \sum_{x_t \in D_T} J (z_t, y_t)\big),
\end{equation}


\noindent where J(.,.) is the cross entropy loss, $y_s$ is the label of source data and $y_t$ is the pseudo label of target data. The classifier $C$ is pre-trained with the supervised source data for obtaining the high confident pseudo labeled target data. It is notable that $C$ can be any type of deep learning network. Furthermore, the classifier assists to achieve the domain invariant latent representation as the domain-specific portion in latent space.



\subsection{Overall Objective}
The overall objective function of our proposed SCGAN can be formulated as follows,
\begin{equation}
L_{total} = \underset{E,C,G,D}{min} (L_G + L_D + \gamma L_c),
\end{equation}

\noindent where $\gamma$ is a hyper parameter of the classifier network. To preserve the category information of the source and target data, the generator and the discriminators are optimized in an adversarial way.

\begin{figure}
\begin{center}
\includegraphics[width=1.0\linewidth]{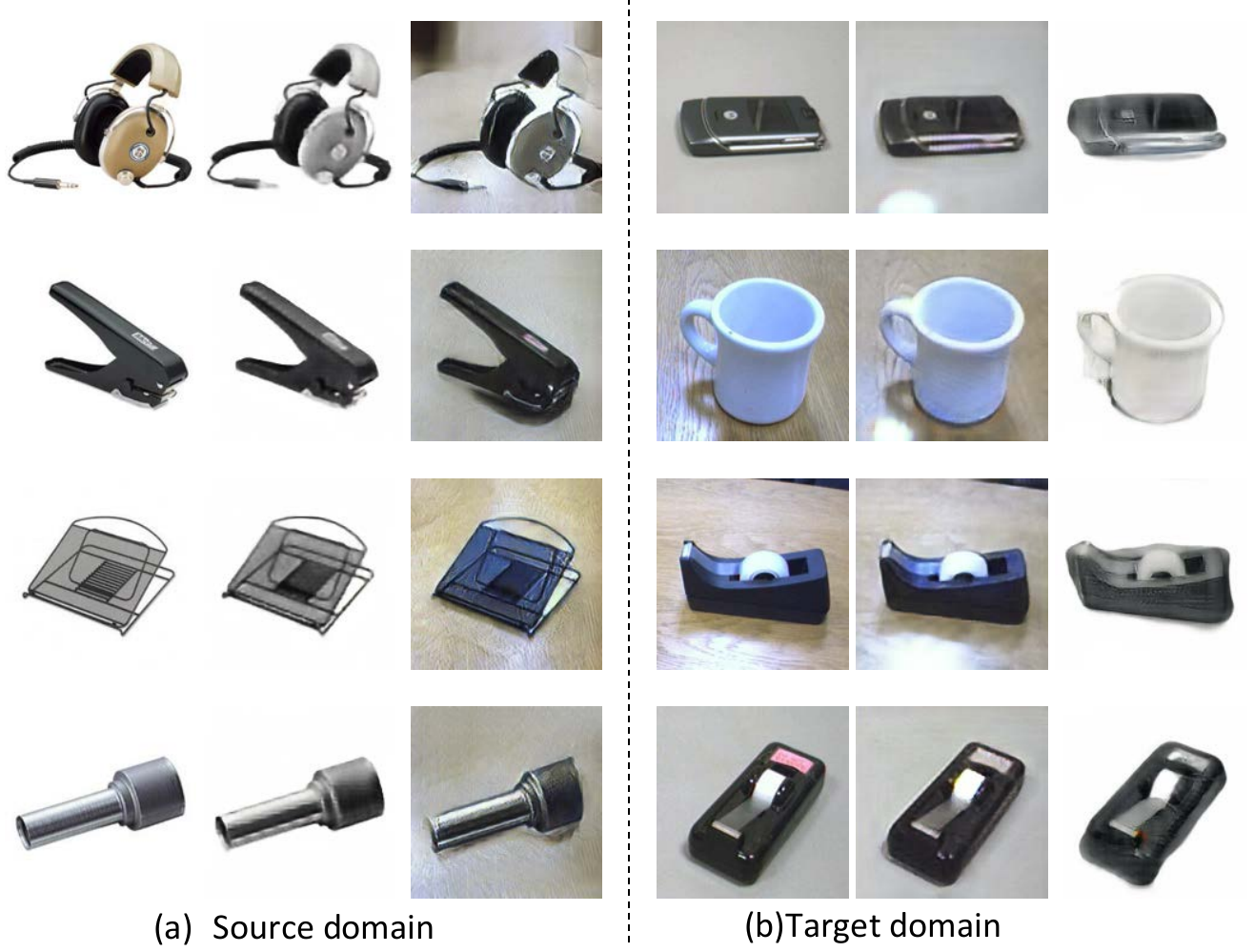}
\end{center}
   \caption{Examples of generated images during Amazon (A) $\rightarrow$ Webcam (W) transfer task. The first and fourth columns represent the source and target domain's real images respectively. The second and third columns represent the generated images from source domain with source and target domains style respectively. The fifth and sixth columns represent the generated images from target domain with target and source domains style respectively.}
\label{fig:generated_image}
\end{figure}

\begin{table*}[h!]
\centering
\caption{Recognition accuracies for cross-domain digit classification tasks. We use the conventional protocol for UDA where source data are labeled, but target data are unlabeled. MNIST $\rightarrow$ USPS indicates MNIST is the source and USPS is the target domain.}

   \resizebox{12cm}{!}{

 \begin{tabular}{|c | c | c | c | c | c |} 


 \hline
 \textbf{Method}  & \textbf{MNIST} $\rightarrow$ \textbf{USPS}  & \textbf{USPS} $\rightarrow$ \textbf{MNIST}  &\textbf{SVHN} $\rightarrow$ \textbf{MNIST} &  \textbf{MNIST} $\rightarrow$ \textbf{SVHN} & \textbf{MNIST} $\rightarrow$ \textbf{MNIST-M} \\   
 \hline
 Source Only &86.2  &75.3  &61.9  &32.3 &69.7 \\
 

 CORAL \cite{sun2016return} &81.7  &-  &63.1  &- &76.9 \\
 
 MMD \cite{7410820} &81.1  & -  &71.1  & - &76.9 \\
 
 DANN \cite{pmlr-v37-ganin15} &85.1  &74.2  &73.9  &35.7 &77.4 \\ 
 
 
 ADDA \cite{8099799} &89.4  &90.1  &76.0  &- &- \\
 
 
 
 ATT \cite{saito2017asymmetric} &-  &-  &86.2  &52.8 &94.2 \\ 
 
 
 CoGAN \cite{NIPS2016_6544} &91.2  &89.1  &-  &- &62.0 \\ 
 
 UNIT \cite{liu2017unsupervised} &95.9  &93.5  &90.5  &- &- \\ 
 
 
 SBADA-GAN \cite{rucvpr} &97.6  &95.0  &76.1  &61.1 &\textbf{99.4} \\ 
 
 GTA \cite{Gen2Adapt} &92.8  &90.8  &92.4  &- &- \\ 
 
 CyCADA \cite{pmlr-v80-hoffman18a} &94.8  &95.7  &88.3  &-  &-\\ 
 
 DupGAN \cite{Hu_2018_CVPR} &96.0  &\textbf{98.8}  &92.5  &62.7 &-\\ 
 
 I2I \cite{murez2018image} &\textbf{98.8} &97.6 &90.1 &- &-\\
 \hline
 \textbf{SCGAN(Ours)} &97.9  &98.7  &\textbf{94.5}  &\textbf{65.8} & 99.2  \\ 
 \hline
 \end{tabular}}
\label{table_digit} 
 
\end{table*}


\begin{table*}[!htbp]
\caption{Image classification accuracies on the Office-31 dataset for deep domain adaptation. For UDA, we use the standard protocol where source data is labeled while target data is unlabeled. $A \rightarrow W$ shows $A$ (Amazon) is the source and $W$ (Webcam) is the target domain.}

\begin{center}
\small\addtolength{\tabcolsep}{10pt}
\resizebox{12cm}{!}{
\begin{tabular}{|c|c|c|c|c|c|c|c|}





\hline

\textbf{Methods} & \textbf{A} $\rightarrow$ \textbf{W} & \textbf{D} $\rightarrow$ \textbf{W} & \textbf{D} $\rightarrow$ \textbf{A} & \textbf{W} $\rightarrow$ \textbf{A}& \textbf{W} $\rightarrow$ \textbf{D} & \textbf{A} $\rightarrow$ \textbf{D} & \textbf{Avg.}\\ [0.5ex] 
\hline

 Resnet-Source Only &68.4 &96.7&62.5 &60.7 &99.3 &68.9 &76.1 \\
 
 DAN \cite{DBLP:conf/icml/LongC0J15}  &80.5  & 97.1 & 63.6 &62.8 &99.6 &78.6  &80.4\\
 RTN \cite{DBLP:conf/nips/LongZ0J16}  &84.5 &96.8 &66.2 &64.8 &99.4 &77.5 &81.6\\
 
  DANN \cite{pmlr-v37-ganin15}  &82.0 &96.9 &68.2 &67.4 &99.1 &79.7 &82.2\\
 ADDA \cite{8099799} &86.2 &96.2 &69.5 &68.9 &98.4 & 77.8 &82.9\\
 
 JAN \cite{DBLP:conf/icml/LongZ0J17}  &85.4 &97.4 &68.6 &70.0 &\textbf{99.8} &84.7 &84.3\\

  MADA \cite{pei2018multi} &90.0 &97.4 &70.3 &66.4 &99.6 & 87.8 &85.2 \\
 
 SimNet \cite{Zhang_2019_CVPR} &88.6 &\textbf{98.2} &73.4 &71.6 & 99.7 & 85.3 &86.2 \\
 
 
 GTA \cite{Gen2Adapt} &89.5 &97.9 &72.8 &71.4 &\textbf{99.8} &87.7 &86.5\\
\hline
\textbf{SCGAN (Ours)}  &\textbf{91.5} &98.1 &\textbf{74.5} &\textbf{73.6} &\textbf{99.8} &\textbf{90.3} &\textbf{88.0}\\
\hline
\end{tabular}}
\end{center}
\label{office31}
\end{table*}

\vspace{-4mm}

\section{Experiments and Results}
\vspace{-2mm}

\subsection{Datasets}
We evaluate our proposed SCGAN on digit and object classification tasks. \textbf{MNIST} \cite{lecun-gradientbased-learning-applied-1998} consists of 10 classes which ranges 0 to 9. \textbf{USPS} \cite{hastie01statisticallearning} dataset is created by taking the digit images which are automatically scanned from United States Postal Service's envelopes. It contains 9,298 images with 10 classes ranges from 0 to 9. \textbf{SVHN} \cite{SVHN2011} is extracted from the street view house number, comprising of more than 600,000 colored images. \textbf{MNIST-M} \cite{pmlr-v37-ganin15} dataset comprises MNIST digits where the background is blended with random color patches. For MNIST $\rightarrow$ USPS, USPS $\rightarrow$ MNIST, SVHN  $\rightarrow$ MNIST, MNIST $\rightarrow$ SVHN and MNIST $\rightarrow$ MNIST-M transfer tasks, we use the same protocol as \cite{Hu_2018_CVPR,Gen2Adapt}. \textbf{Office-31}  \cite{Saenko:2010:AVC:1888089.1888106} dataset consists of three domains: Amazon (A), Webcam (W) and DSLR (D). It has 31 different classes. We adopt the same protocol as \cite{Gen2Adapt,Zhang_2019_CVPR} for A $\rightarrow$ W, D $\rightarrow$ W, D $\rightarrow$ A, W $\rightarrow$ A, W $\rightarrow$ D and A $\rightarrow$ D transfer tasks.

\begin{table*}[h!]
\centering
\caption{Ablation study of our proposed approach on unsupervised domain adaptation.}
 \resizebox{12cm}{!}{

 \begin{tabular}{|c | c | c | c | c | c |} 


 \hline
 \textbf{Method}  & \textbf{SVHN} $\rightarrow$ \textbf{MNIST}  & \textbf{MNIST} $\rightarrow$ \textbf{SVHN}  &\textbf{A} $\rightarrow$ \textbf{D} &  \textbf{D} $\rightarrow$ \textbf{A} \\   
 \hline
SCGAN (without semantic loss) &92.8 &63.2 &88.1 &72.9 \\
SCGAN (with semantic loss) &94.5 &65.8 &90.3 &74.5 \\
 \hline
 \end{tabular}}
\label{table_Ablation} 
\end{table*}

\vspace{-4mm}

\section{Visualization}
\vspace{-4mm}

\begin{figure*}[h]
\begin{center}
\includegraphics[width=1.0\linewidth]{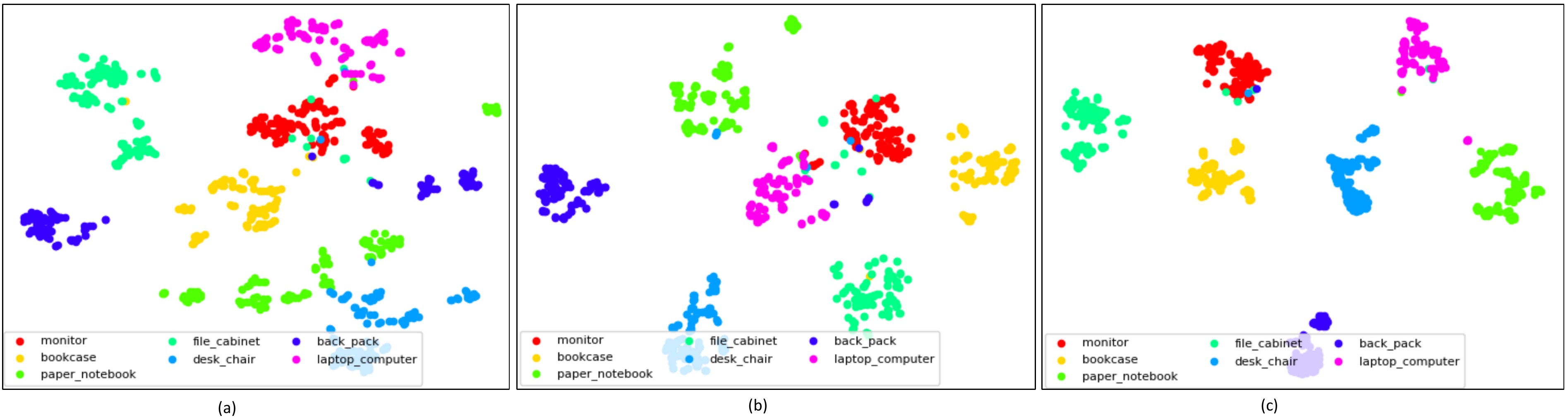}
\end{center}
   \caption{t-SNE visualization of the activations of (a) Resnet without adaptation (b) SCGAN-without semantic consistent loss and (c) SCGAN. }
\label{fig:tsne}
\end{figure*}

\subsection{Experiments}
We evaluate our proposed SCGAN in the context of image classification tasks where labeled source data and unlabeled target data is used during training. The performance is reported in terms of classification accuracy. We compare our SCGAN approach for UDA with some state-of-the-art methods. For digit classification tasks, we compare our approach with CORAL \cite{sun2016return}, MMD \cite{7410820}, DANN \cite{pmlr-v37-ganin15}, ADDA \cite{8099799}, ATT \cite{saito2017asymmetric}, CoGAN \cite{NIPS2016_6544}, UNIT \cite{liu2017unsupervised}, SBADA-GAN \cite{rucvpr}, GTA \cite{Gen2Adapt}, CyCADA \cite{pmlr-v80-hoffman18a}, DupGAN \cite{Hu_2018_CVPR} and I2I \cite{murez2018image}. For object classification tasks on Office-31 dataset, we compare our approach with DAN \cite{DBLP:conf/icml/LongC0J15}, RTN \cite{DBLP:conf/nips/LongZ0J16}, DANN \cite{pmlr-v37-ganin15}, ADDA \cite{8099799}, JAN \cite{DBLP:conf/icml/LongZ0J17},  MADA \cite{pei2018multi}, SimNet \cite{Zhang_2019_CVPR} and GTA \cite{Gen2Adapt}. For more in-depth comparison we also compare the results with a source only method where the same architecture is used with the labeled source data.

\vspace{-4mm}
\subsection{Implementation Details}

In the experiments of digit classification tasks, we employ the architecture of convolutional neural network used in  \cite{saito2017asymmetric} and  \cite{liu2017unsupervised}. For MNIST $\rightarrow$ USPS, USPS $\rightarrow$ MNIST, SVHN $\rightarrow$ MNIST transfer tasks, we follow the same architecture of  \cite{liu2017unsupervised} and for MNIST $\rightarrow$ SVHN, MNIST $\rightarrow$ MNISTM transfer tasks we follow the same architecture of  \cite{saito2017asymmetric}. The impact of pseudo-labels in a neural network was studied in \cite{lee2013pseudo}. They claimed that training a classifier with pseudo-labels has the same effect as entropy regularization, resulting in low-density class separation. Inspired by \cite{lee2013pseudo}, the classifier $C$ is pre-trained with only source domain images to achieve mostly accurate and  high-confident pseudo-labeled target domain samples. The pseudo label with softmax score greater than a threshold is chosen to train the predicting model. In MNIST $\rightarrow$ SVHN, we set the pseudo labeling threshold value to 0.95. We picked it to $0.9$ in in other transfer tasks. For optimization, we use Momentum SGD and set the momentum to 0.9. The learning rate is calculated on validation splits and uses either 0.01 or 0.05. For MNIST $\rightarrow$ USPS, USPS $\rightarrow$ MNIST, and SVHN $\rightarrow$ MNIST transfer tasks, the value of $\alpha$, $\beta$ and $\gamma$ are set as 10.0, 1.0 and 0.2 respectively, and for MNIST $\rightarrow$ SVHN, MNIST $\rightarrow$ MNIST-M transfer tasks and all other object recognition tasks, the value of $\alpha$, $\beta$ and $\gamma$ are set as 1.0, 1.0 and 1.0 respectively.


\vspace{-1mm}
\subsection{Results}
\vspace{-1mm}
In this section, we evaluate the efficacy of the proposed method on two classification tasks: digit classification and image classification. The proposed method is compared with state-of-the-art methods as shown in Table \ref{table_digit} and Table \ref{office31}. 

Table \ref{table_digit} shows the recognition accuracies for digit experiments on four benchmark datasets: MNIST, USPS, SVHN and MNIST-M. For fair comparison, we follow the same settings as followed by the state-of-the-art methods. We adopt five UDA settings: MNIST $\rightarrow$ USPS, USPS $\rightarrow$ MNIST, SVHN $\rightarrow$ MNIST, MNIST $\rightarrow$ SVHN and MNIST $\rightarrow$ MNIST-M to evaluate the proposed SCGAN. In MNIST $\rightarrow$ USPS, SVHN $\rightarrow$ MNIST, and MNIST $\rightarrow$ SVHN settings, the previous state-of-the-art approach DUPGAN achieved 96.0\%, 92.5\% and 62.7\% recognition accuracies, however, proposed SCGAN achieves a 1.9\%, 2.0\% and 3.1\% increase over DUPGAN, indicating a substantial improvement in performance with preserving semantic features across domains. Although in USPS $\rightarrow$ MNIST and  MNIST $\rightarrow$ MNIST-M experiments, DUPGAN and SBADA-GAN outperforms our method by 0.1\% and 0.2\% recognition accuraices, which is a slight increase in performance. In particular, we achieve good performance improvement, i.e., up to +3 \% points in one of the most challenging tasks, MNIST $\rightarrow$ SVHN. To further evaluate the proposed method, our deep network is trained with the label information of the source domain only, referred to as Source Only when the target domain is not available.

For object classification, Office-31 dataset is used to evaluate our SCGAN. We evaluate six transfer tasks A $\rightarrow$ W, D $\rightarrow$ W, D $\rightarrow$ A, W $\rightarrow$ A, W $\rightarrow$ D and A $\rightarrow$ D on this dataset. We adopt the conventional unsupervised protocol for DA where labeled source and unlabeled target data are used. Our SCGAN is compared with state-of-the-art methods in Table \ref{office31}. Our method achieves 88.0\% average accuracy and the previous state-of-the-art method \cite{Gen2Adapt} acheived 86.5\%. As shown in Table \ref{office31}, proposed SCGAN outperforms all the compared methods, especially in all challenging transfer tasks:  A $\rightarrow$ W,  D $\rightarrow$ A,  W $\rightarrow$ A, and A $\rightarrow$ D.



\vspace{-1mm}
\subsection{Ablation Study}
\vspace{-1mm}
We study the effect of adding semantic consistent loss to clarify the role capturing semantic information of our proposed method. We train the network without semantic consistent loss and with semantic consistent loss for SVHN $\rightarrow$ MNIST, MNIST $\rightarrow$ SVHN, A$\rightarrow$ D and D$\rightarrow$ A transfer tasks as shown in Table \ref{table_Ablation} to justify the effectiveness of our proposed model. According to Table \ref{table_Ablation}, for SVHN $\rightarrow$ MNIST and MNIST $\rightarrow$ SVHN, without semantic consistent loss, the performance drops by 1.7\% and 2.6\% respectively than SCGAN with semantic consistent loss. For object classification task, we perform experiment on A$\rightarrow$ D and D$\rightarrow$ A transfer tasks without the semantic loss. However, the performance has 2.2\% and 1.6\% decrease than considering the model with the semantic consistent loss.


We evaluate the effectiveness of our proposed approach and perform a t-SNE visualization of the learned embeddings. Figure \ref{fig:tsne} shows a t-SNE visualization of the learned embeddings on the $A \rightarrow W$ transfer task. Here we have considered 7 classes for better visualization. We observe that the clusters formed by SCGAN separate classes while mixing source and target domains much more effectively than the method without domain adaptation and SCGAN-without semantic loss. Clusters generated by our proposed model are observed to be able to distinguish classes far more effectively than SCGAN without semantic consistent loss and without domain adaptation approach. Thus we can say that SCGAN and SCGAN without semantic consistent loss reduce the discrepancy between the source and target data in the latent embedding space more than the method without domain adaptation. From Figure 3, we can observe that SCGAN less wrongly clustered points in the red clusters (class: monitor) than SCGAN without semantic consistent loss and without domain adaptation approaches.
\vspace{-3mm}

\section{Conclusion}
\vspace{-4mm}

In this paper, we propose an end-to-end semantic consistent generative adversarial network for unsupervised domain adaptation. We aim to capture the semantic information across the source and target domains in the feature level during the domain alteration for minimizing the discrepancy among domains and utilize pseudo labels assigned to target data for boosting the performance. The domain alteration is achieved with one encoder, one generator, and two discriminators for the source and target domains. A classifier which is assembled on the encoder is utilized to predict the categories of the source and target data. We demonstrate the performance of our approach for unsupervised domain adaptation on digit and object detection tasks, and show that it offers superior or competitive performance compared to the state-of-the-art methods that have been reported to date.

\section*{Acknowledgements}
The research presented in this paper was supported by Australian Research Council (ARC) Discovery Project Grant
DP170100632.

\bibliography{mybibfile}

\end{document}